\documentclass[runningheads]{llncs}
\usepackage{amsmath,amssymb}
\usepackage{graphicx}
\usepackage[caption=false]{subfig}
\usepackage{microtype}
\usepackage{algorithm}
\usepackage{algpseudocode}
\usepackage{siunitx}
\usepackage{cite}
\usepackage{pgfplotstable}

\pgfplotsset{compat=1.18}

\newcommand{\best}[1]{\textbf{#1}}
\newcommand{\second}[1]{\underline{#1}}

\newcommand{\mysubsection}[1]{\medskip\noindent\textbf{#1}}

\usepackage{xcolor}
\newcommand{\legenditem}[2]{\textcolor{#1}{\rule{1.6em}{0.8ex}}~#2}

\definecolor{mygreen}{HTML}{008000}
\definecolor{lightgreen}{HTML}{90EE90}
\definecolor{myred}{HTML}{FF0000}
\definecolor{lightcoral}{HTML}{F08080}
\definecolor{myblue}{HTML}{0000FF}
\definecolor{mypurple}{HTML}{800080}
\definecolor{mypink}{HTML}{FFC0CB}
\definecolor{mybrown}{HTML}{A52A2A}
\definecolor{myorange}{HTML}{FFA500}
\definecolor{myyellow}{HTML}{E6C300}
\definecolor{mycyan}{HTML}{00FFFF}

\usepackage{booktabs, threeparttable, tabularx, makecell}

\title{On Improving Deep Active Learning with Formal Verification}
\author{Jonathan Spiegelman\inst{1} \and Guy Amir\inst{2} \and Guy Katz\inst{1}}

\institute{
  The Hebrew University of Jerusalem\\
  \email{\{yhonatan.spigelman, g.katz\}@mail.huji.ac.il}
  \vspace{0.1cm}
  \and
  Cornell University\\
  \email{gda42@cornell.edu}
}

\begin{document}
\maketitle

\begin{abstract}
  \vspace{-0.5cm} 
  Deep Active Learning (DAL) aims to reduce labeling
  costs in neural-network training by prioritizing the most
  informative unlabeled samples for annotation. Beyond selecting which
  samples to label, several DAL approaches further enhance data
  efficiency by augmenting the training set with synthetic inputs that
  do not require additional manual labeling. In this work, we
  investigate how augmenting the training data with adversarial inputs
  that violate robustness constraints can improve DAL performance. We
  show that adversarial examples generated via formal verification
  contribute substantially more than those produced by standard,
  gradient-based attacks.  We apply this extension to multiple modern
  DAL techniques, as well as to a new technique that we propose, and
  show that it yields significant improvements in model generalization
  across standard benchmarks.  
\end{abstract}

\section{Introduction}
\label{sec:introduction} 
Deep Neural Networks (DNNs) have achieved remarkable success in a wide
range of
applications~\cite{KrSuHi12,DeChLeTo2019,SiHuMaGuSiVaScAnPaLaDi16}. However,
training these models in a supervised setting typically requires vast
amounts of labeled
data~\cite{SuShSiGu17,HeNaArDiJuKiPaYaZh17}. Acquiring labels may be
costly, e.g., when highly skilled human experts are needed.  Deep
Active Learning (DAL) aims to minimize this expense by labeling the
most informative samples under a limited labeling
budget~\cite{Se09}. In general, sample selection can be driven either
by the model's uncertainty---prioritizing examples it is least
confident about---or by diversity considerations that encourage
choosing samples that better represent the broader data
distribution~\cite{ReXiChHuLiGuChWa21}. 

In parallel, the formal methods community has developed verification
tools that take a network and a specification, and either formally
certify that the specification holds or return a counterexample
(input) that violates it. For instance, verifiers can prove a
network's robustness to perturbations around a given input, or produce
perturbations that cause an error~\cite{HuKwWaWu17,GeMiDrTsChVe18,KaBaDiJuKo17Reluplex,LeYeKa23}. 
Unlike one-off gradient-based attacks, a verifier can systematically explore
a bounded input region to find \emph{multiple}, \emph{distinct}
adversarial examples (or certify that none exist)---and such a diverse 
set can be highly beneficial for DAL, as it enriches and amplifies the 
training set once informative samples are selected.

In this paper, we explore the integration of adversarial inputs
produced by formal verification as a form of training-set
augmentation. Namely, for each newly labeled point, we employ a
verifier to explore its neighborhood within a bounded radius, finding
several adversarial inputs. These inputs are assigned the oracle's
label of the original sample, and added to the training set at no
additional labeling cost. As we show, this addition helps the learning
process converge to superior results more quickly.

To assess our approach, we evaluate it across a variety of standard
benchmarks (MNIST, fashionMNIST, CIFAR-10); and also across multiple DAL methods, ranging
from lightweight random sampling to more sophisticated approaches,
including a novel approach that we propose here. For each of these
approaches, we add a verification phase after point selection is
carried out. All across the board, we see that this extra step affords
significant improvements. 
To our knowledge, ours is the first work to incorporate adversarial
examples generated via formal verification into the active-learning
loop, enabling a systematic data augmentation that amplifies each
human label.

\section{Background}
\label{sec:background} 
Deep neural networks compose affine transformations with nonlinear activations (e.g., ReLUs) to learn high-capacity representations~\cite{LeBeHi15}. Despite their success, it has been shown that small, often imperceptible input perturbations can dramatically alter a network's prediction~\cite{SzZaSuBrErGoFe13}, revealing a persistent robustness gap. Gradient-based attacks such as FGSM~\cite{GoShSz14} and PGD~\cite{MaMaScTsVl17} vividly demonstrate this fragility by finding small perturbations that cause misclassification. These attacks have become standard tools for probing model robustness. 

Formal verification tackles the robustness problem from a different angle. Given a network and a formal specification (e.g., ``the class of input $\mathbf{x}$ remains $\mathbf{y}$ within an $\ell_p$-ball''), a verifier will either prove the property holds or produce a counterexample input that violates it~\cite{HuKwWaWu17,GeMiDrTsChVe18,CoAmRoSaKaFo24,ElCoKa23}. Compared to a single-step attack, verification can systematically
explore a bounded neighborhood of an input, yielding a stronger
coverage of potential adversarial examples, albeit at a higher computational cost~\cite{IsZoBaKa23}. 

The field of Deep Active Learning (DAL) aims to reduce labeling effort
by querying only the most informative samples from an unlabeled
pool~\cite{Se09}. A typical DAL process starts by
training a model on a small initial labeled set $\mathcal{L}$, often a
random subset of the unlabeled dataset $\mathcal{U}$. It then proceeds
iteratively: in each round, the model evaluates the remaining
unlabeled instances according to some criterion, selects a batch of
top-scoring instances, and sends it to an oracle for labeling. The
newly labeled samples are then added to $\mathcal{L}$, and the model
is retrained. This loop repeats until a labeling budget is
exhausted.

DAL query strategies are often categorized along an
uncertainty-diversity spectrum~\cite{ReXiChHuLiGuChWa21}. Uncertainty
sampling targets points on which the model is most unsure (e.g.,
lowest confidence~\cite{CuMc05} or highest
entropy~\cite{Sh48}), while diversity-based methods aim
to efficiently cover the data distribution (e.g.,
CoreSet~\cite{SeSa18}). Hybrid methods, such as
BADGE~\cite{AsZhKrLaAg19}, combine both criteria, selecting batches of
examples aimed at simultaneously improving diversity and reducing
uncertainty. In addition to query strategies, some DAL approaches
incorporate \emph{data augmentation}---the adding of synthetic inputs
to the training set---to improve model generalization without extra
labeling costs~\cite{MoAnEcBo24,ChZhWaChLi22}, using
various methods such as GANs~\cite{NiOk19}. A compelling
approach is to leverage adversarial examples: around each labeled
sample, one can generate minimally perturbed variants that the current
model misclassifies, and add them to the training set with the same
label as the original sample~\cite{LiLiChWaDoHu23}.

One principled selection criterion, especially for classification
tasks, is to query samples near the model's decision boundary---i.e.,
those with the \emph{smallest margins}~\cite{ToKo01, BaBrZh07}. In the DAL
taxonomy, margin-based selection is considered an uncertainty-driven
strategy: a small margin implies the model is indecisive about the
sample's label. Since exact margin computation in DNNs is intractable,
the size of the smallest adversarial perturbation serves as a
practical proxy for this value. DFAL~\cite{DuPr18},
for instance, uses the DeepFool attack~\cite{MoSeFaFr16} to
estimate each sample's perturbation margin and selects the
lowest-margin candidates; then, it augments the training set by
including each sample's corresponding adversarial example, labeled identically to its source. This way, DFAL combines margin-based selection with data augmentation. While effective, DFAL uses only a single adversarial example per query, offering a limited exploration of each sample's local decision boundary.

\section{Our Approach: Augmentation via Verification}
\label{sec:fvaal}
\vspace{-1mm}
We propose to leverage a formal verification engine to create a set of multiple, diverse adversarial examples for each queried input, thereby amplifying the value of each oracle query. In particular, we use a verifier to systematically probe the model's decision boundaries around each selected sample and find additional adversarial inputs that cause the model's prediction to flip to a different class.

Formally, given an input $\mathbf{x}$ with a current model label
$\hat{y} = \arg\max{f(\mathbf{x})}$, the specification we want to
check is whether there exists an input within
$[\mathbf{x}-\epsilon,\mathbf{x}+\epsilon]$ where the property
$f(\mathbf{x})_{i}>f(\mathbf{x})_{\hat{y}}$ holds for any class $i\ne
\hat{y}$. In addition, we consider the adversarial inputs already
found, and exclude a small region around each of them. For example, if
a known adversarial input $\mathbf{x}'$ has $\mathbf{x}'_1 = 0.5$, we
can constrain the next query with $(\mathbf{x}_1 < 0.4999 \vee
\mathbf{x}_1 > 0.5001)$, to avoid re-discovering that exact
counterexample. We repeatedly invoke the verifier with the updated
constraints, harvesting up to $k$ distinct adversarial examples
${\mathbf{x}^{(1)},\mathbf{x}^{(2)},\ldots\mathbf{x}^{(k)}}$ for the
same input $\mathbf{x}$. If the verifier fails to find such inputs,
the $\epsilon$ value is slightly increased and the verifier is run again with the same query. The process runs until a time limit is reached. All samples are eventually assigned the same label obtained from the oracle for the original input. This way, a single query point $\mathbf{x}$ contributes a local cloud of high-value samples. 

There exist lightweight approaches, besides verification, to identify
adversarial examples. To measure the usefulness of verification in
this context, we compare our approach to the Fast Gradient Sign Method
(FGSM)~\cite{GoShSz14}, a gradient attack which generates an
adversarial example by applying a perturbation in the direction of the
input gradient's sign. Namely, given an input $\mathbf{x}$ and a loss
function $L$, the perturbation is
$\epsilon \cdot \text{sign}(\nabla_{\mathbf{x}} L)$ where $\epsilon$
is a hyperparameter. While FGSM can produce multiple adversarial
inputs by running it with various $\epsilon$ values, it is confined to
perturbations along a single gradient direction. In contrast,
verification can reach off-gradient failure regions and produce
multiple counterexamples that expose distinct modes of
misbehavior---producing a more diverse set of adversarial inputs.

\section{Evaluation}
\label{sec:eval}

We evaluate our data augmentation method on supervised image
classification tasks, using classic benchmarks that are often
considered in active learning papers: MNIST~\cite{Le98},
fashionMNIST~\cite{XiRaVo17} and
CIFAR-10~\cite{KrHi09}, each of which consists of images
with 10 possible classes. Additional technical details about the models, the
training process and the backend verifier can be found in Appendix~\ref{app:tech-details}, and
our code and benchmarks are available online~\cite{OurCode}. We
use data augmentation on several DAL methods from the literature:

\begin{itemize}
    \item \texttt{DFAL} (DeepFool Active Learning): An adversarial query strategy that uses the DeepFool attack to approximate the model's decision margin~\cite{DuPr18}. It selects unlabeled samples requiring the smallest adversarial perturbation and queries them for labeling. Each chosen sample's adversarial instance is also labeled with the same class and added the training set.
    \item \texttt{BADGE}: A DAL method that selects points with large
      output-layer gradient magnitudes while ensuring diversity in
      their feature representations~\cite{AsZhKrLaAg19}. \texttt{BADGE} computes a ``gradient embedding'' for each unlabeled instance and then uses the $k$-means++ clustering algorithm~\cite{ArVa07} to choose a batch of $k$ points whose gradient vectors are both high in norm and diverse in direction.
    \item \texttt{Random}: A simple baseline that selects unlabeled examples uniformly at random from the unlabeled set. This method uses no information from the model, serving as a lower-bound benchmark.
\end{itemize}

In addition to these existing DAL methods, we also propose
\texttt{FVAAL}---a margin-based DAL strategy that leverages
adversarial perturbations to estimate an unlabeled input's distance to
the model's decision boundary. \texttt{FVAAL} builds on FGSM to
estimate a sample's margin---while in standard FGSM a fixed $\epsilon$
value is chosen to induce misclassification, we \emph{iteratively}
search for the smallest perturbation that causes the model's
prediction to flip, using a binary search over $\epsilon$, which is
stopped when the search interval reaches a sufficiently low value. The
smallest successful $\epsilon$ value, denoted $\epsilon^*$, serves as
a proxy for the input's distance to the decision boundary (i.e., the
margin). Unlabeled samples can then be ranked by $\epsilon^*$, with
smaller values indicating points that are closer to the boundary and
thus more informative under margin-based criteria. The samples with
smallest $\epsilon^*$ values are sent for labeling, and for each
sample---its generated adversarial input is also added to the training
set, with the same label as the one obtained from the oracle for the
original image. On top of this selection mechanism, we use a formal
verifier to enhance the labeled samples using verification-generated
inputs, as explained above.  \texttt{FVAAL} uses a
simple-yet-efficient approximation of the margin for each
sample. Unlike more expensive, multi-step attacks such as
DeepFool~\cite{MoSeFaFr16}, which iteratively refine perturbations to
find minimal distortion, this method requires only a single
calculation of the gradient, and a simple forward pass each iteration
for checking misclassification. This efficiency is crucial in DAL
scenarios, where thousands of unlabeled candidates may need to be
assessed. The full pseudo-code of \texttt{FVAAL} appears in
Appendix~\ref{app:algorithms}.

\begin{figure}[t]
  \centering
  \subfloat[MNIST]{
    \includegraphics[width=0.31\textwidth]{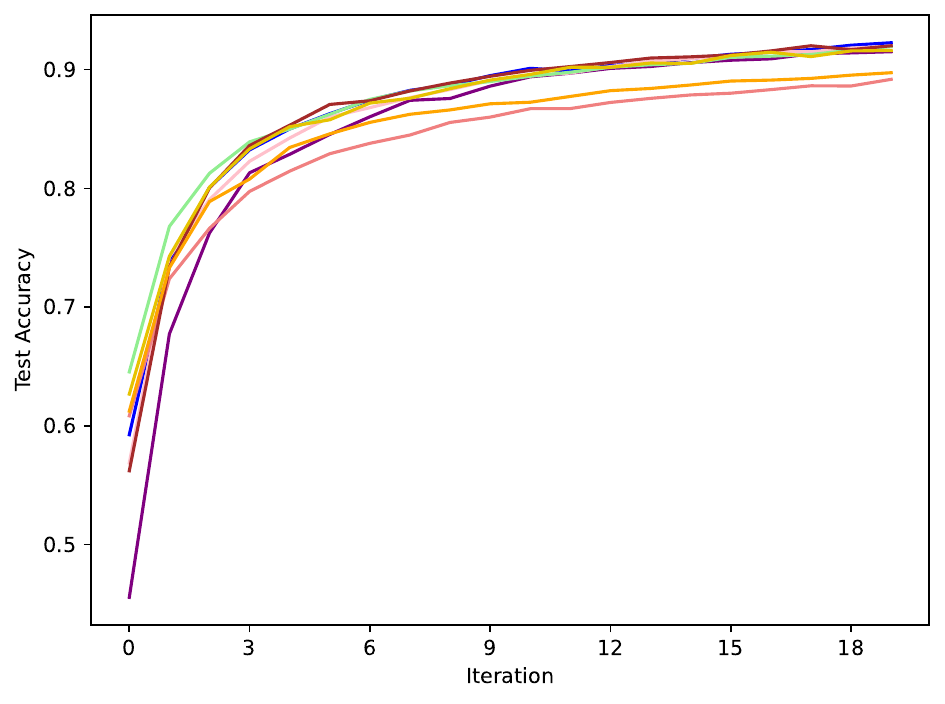}
  }
  \subfloat[fashionMNIST]{
    \includegraphics[width=0.31\textwidth]{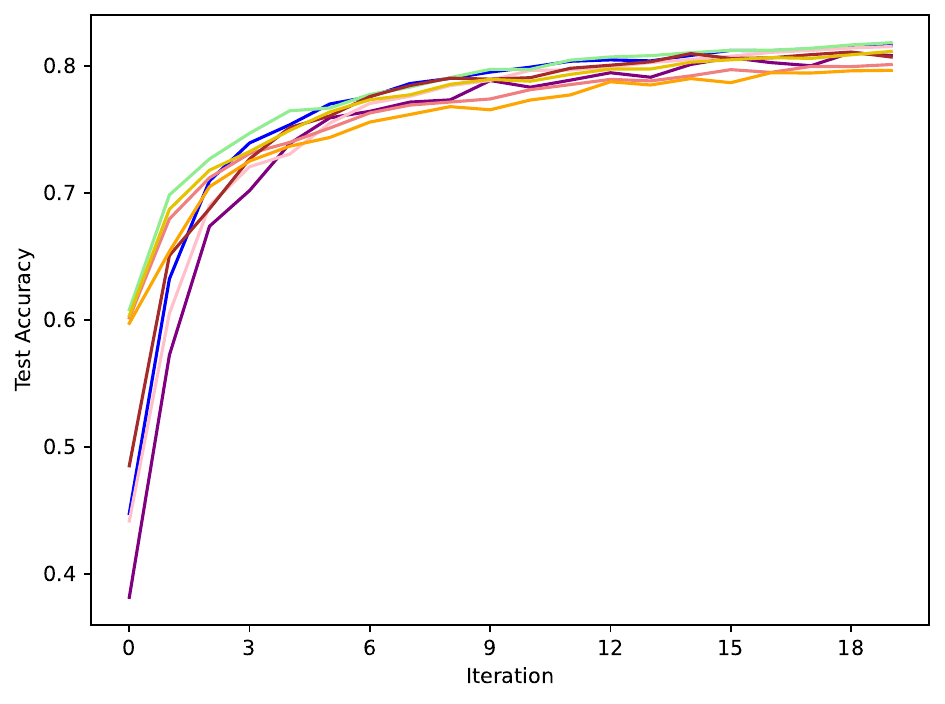}
  }
  \subfloat[CIFAR-10]{
    \includegraphics[width=0.31\textwidth]{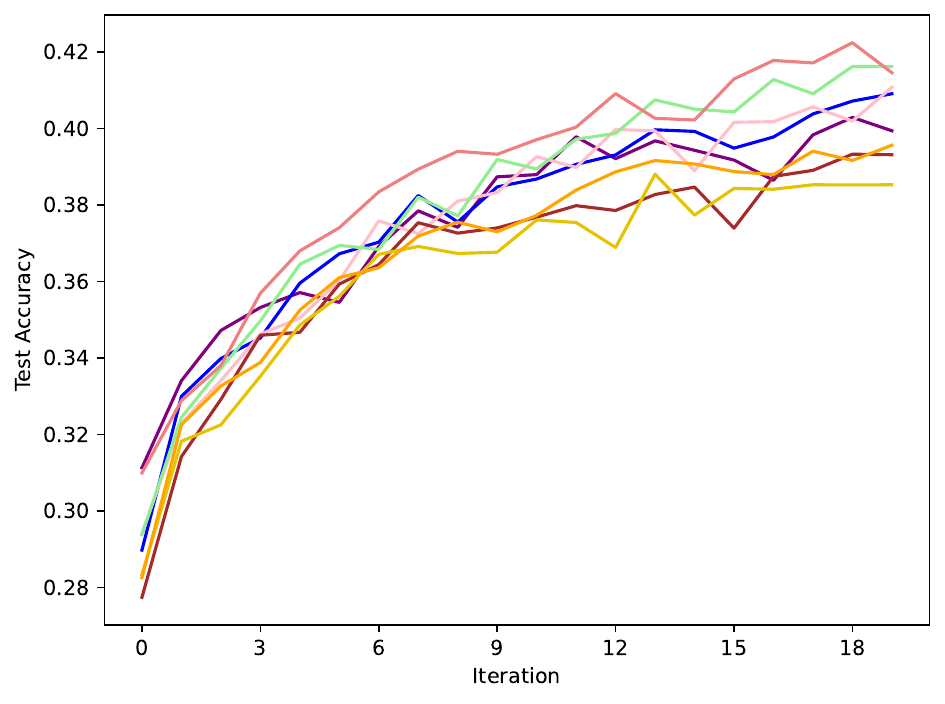}
  }

  \begin{minipage}{0.95\linewidth}\centering\small
  \legenditem{lightgreen}{BADGE+FV-Adv}\quad
  \legenditem{myyellow}{BADGE+FGSM-Adv}\quad
  \legenditem{mypurple}{DFAL}\quad
  \legenditem{mypink}{DFAL+FV-Adv}\quad
  \legenditem{mybrown}{DFAL+FGSM-Adv}\quad
  \legenditem{myblue}{FVAAL}\quad
  \legenditem{lightcoral}{Random+FV-Adv}\quad
  \legenditem{myorange}{Random+FGSM-Adv}
\end{minipage}
  \caption{Test accuracy vs. number of iterations on the three benchmarks with adversarial inputs}
  \label{fig:accuracy-adv}
  \vspace{-2mm}
\end{figure}

For each method we compare two variants:
(a) for each labeled sample we generate up to 10 adversarial examples
created by the Marabou formal verifier~\cite{KaHuIbHuLaLiShThWuZeDiKoBa19,WuIsZeTaDaKoReAmJuBaHuLaWuZhKoKaBa24} and add
them to the batch each iteration (``\texttt{+FV-Adv}''); (b) for each
sample we produce up to 10 adversarial inputs using FGSM with a range
of $\epsilon$ values between 0.05 and 0.1
(``\texttt{+FGSM-Adv}''). For \texttt{DFAL}, which natively generates a
single adversarial input per sample, only 9 additional adversarial
inputs were generated by Marabou/FGSM. For comparison, we also added
the original \texttt{DFAL} method, adding only a single adversarial inputs. 
Fig.~\ref{fig:accuracy-adv} depicts, for each of the benchmarks, the test-set classification 
accuracy as a function of the number of DAL iterations. Observing the MNIST and fashionMNIST benchmarks, it appears that using
either approach for generating adversarial inputs produces similar
results---with some gains for formal verification, e.g., when
using \texttt{BADGE} on MNIST or \texttt{Random} on fashionMNIST. We can 
also note \texttt{FVAAL} is one of the top 3 methods in almost every iteration. 
On CIFAR-10, \texttt{+FV-Adv} gives a better learning curve in all baselines. 
Interestingly, variants of the \texttt{Random} method outperform most of the other 
variants on CIFAR-10, with \texttt{Random+FV-Adv} being the best one by a large margin. 
These results imply that incorporating verification as a data augmentation
technique improves the learning curve, and does so better than FGSM. In 
appendix~\ref{app:selection-criterion} we conduct another experiment to 
evaluate \texttt{FVAAL}'s selection critierion, and show it is comparable 
to the other baselines.  

To further analyze labeling efficiency, we use the Area Under the
Budget Curve (AUBC) as a measure for each method's average ``gain''
over each iteration: a higher AUBC indicates better aggregate accuracy
over the DAL process. The complete results appear in
Table~\ref{tab:aubc}. In most cases, verification-based 
inputs achieve a higher AUBC compared to FGSM. Specifically, \texttt{BADGE+FV-Adv} 
has the best AUBC on MNIST and fashionMNIST, and gets second place on CIFAR-10, 
slightly below \texttt{Random+FV-Adv}. \texttt{FVAAL} consistently 
gets the third highest AUBC on all benchmarks, and is second place on 
average. These results confirm that adding verification-generated inputs 
not only improves accuracy measures, but also provides a favorable 
learning curve overall.

\begin{table}[t]
  \centering
  \begin{threeparttable}
  \caption{AUBC across all benchmarks.}
  \label{tab:aubc}
  \setlength{\tabcolsep}{7pt}
  \begin{tabular}{
      l
      S[table-format=1.3(3)]
      S[table-format=1.3(3)]
      S[table-format=1.3(3)]
      S[table-format=1.3(3)]
      S[table-format=1.3(3)]
  }
    \toprule
    {Method} & {MNIST} & {FashionMNIST} & {CIFAR-10} & {Average} \\
    \midrule
    BADGE & {0.8250} & {0.7294} & {0.3723} & {0.6423} \\
    BADGE+FGSM-Adv & {0.8708} & \second{0.7729} & {0.3637} & {0.6691} \\
    BADGE+FV-Adv & \best{0.8738} & \best{0.7812} & \second{0.3820} & \best{0.6790} \\
    DFAL & {0.8554} & {0.7533} & {0.3767} & {0.6618} \\ 
    DFAL+FGSM-Adv & \second{0.8720} & {0.7681} & {0.3664} & {0.6688} \\
    DFAL+FV-Adv & {0.8680} & {0.7626} & {0.3768} & {0.6691} \\ 
    DFAL-No-Adv & {0.8356} & {0.7163} & {0.3664} & {0.6394} \\
    FVAAL & {0.8719} & {0.7712} & {0.3777} & \second{0.6736} \\ 
    FVAAL-No-Adv & {0.8359} & {0.7155} & {0.3680} & {0.6398} \\ 
    Random & {0.8079} & {0.7203} & {0.3706} & {0.6329} \\
    Random+FGSM-Adv & {0.8522} & {0.7575} & {0.3697} & {0.6598} \\
    Random+FV-Adv & {0.8408} & {0.7640} & \best{0.3878} & {0.6642} \\
    \bottomrule
  \end{tabular}
  \begin{tablenotes}[flushleft]
    \footnotesize
    \item Bold = best per benchmark; \underline{underline} = second best.
    \item AUBC is the area under accuracy-budget curve, normalized to $[0,1]$.
  \end{tablenotes}
  \end{threeparttable}
\end{table}

To inspect the diversity of adversarial inputs, in Appendix~\ref{app:diversity} 
we compute all pairwise distances between samples' representations in the model's feature space. 
We show that in all benchmarks and methods, adversarial inputs generated by formal verification provide 
a significantly more diverse set than ones produced by FGSM. We
speculate that this is the reason for the superior performance
demonstrated in Table~\ref{tab:aubc}.

\section{Conclusion}
\label{sec:conclusion}
We presented a data augmentation method for DAL based on adversarial inputs generated 
by a formal verifier, turning each sample into a small, diverse learning
set. We also introduced \texttt{FVAAL}, a DAL method that couples 
fast margin scoring with verification-generated counterexamples. Experiments on standard 
datasets show efficient learning curves and strong AUBC for our proposed augmentation technique. 
The results highlight the potential of formal-verification-based
adversarial augmentation in reducing labeling costs. Future work will
explore the use of additional verifiers, dynamic learning budgets, and
additional datasets.

\mysubsection{Acknowledgments.}
This work was partially funded by the European Union
(ERC, VeriDeL, 101112713). Views and opinions expressed
are however those of the author(s) only and do not necessarily reflect those of the European Union or the European
Research Council Executive Agency. Neither the European
Union nor the granting authority can be held responsible for
them. This research was additionally supported by a grant
from the Israeli Science Foundation (grant number 558/24).

\vspace{-0.1cm}
\bibliographystyle{splncs04}
\bibliography{references}

\newpage

\appendix

\section{Technical Details}
\label{app:tech-details}
\mysubsection{Model Architectures and Training.} We use simple neural network architectures in our experiments. For MNIST and fashionMNIST, we used fully connected networks with one hidden layer consisting of 32 neurons. For CIFAR-10, a similar architecture was chosen, with a hidden layer of 128 neurons. All models used ReLU activation functions in their hidden layers and produced logits. Models were trained using the \texttt{Adam} optimizer with a learning rate of 0.001. We train for 10 epochs in each active learning round, using a mini-batch size of 32.

\mysubsection{Active Learning Protocol.} Each iteration, a subset of 10,000 samples is randomly selected from the unlabeled pool and the selection criterion is run on this sub-pool. For MNIST and fashionMNIST, 50 samples were added each iteration, while in CIFAR-10 1000 samples were added. All models were trained on an initial labeled set of uniformly-selected examples before starting the selection process. The initial labeled set had the same size as the per-iteration batch. We run 20 DAL cycles for each experiment. Each experiment was repeated 5 times and the reported results are averaged over the experiments.

\mysubsection{The Marabou Verifier.} For our evaluation we used the
Marabou DNN verifier as a backend.  Marabou is a modern solver, which
integrates deduction-based and search-based verification
techniques~\cite{WuIsZeTaDaKoReAmJuBaHuLaWuZhKoKaBa24}. Additionally, Marabou
applies abstraction-refinement techniques~\cite{ElCoKa23}, and has
proof-production capabilities~\cite{IsReWuBaKa26, ElIsKaLaWu25,
  BaElLaAlKa25, DeIsKoStPaKa25}. Marabou has been applied to a variety
of tasks, ranging from robotic systems~\cite{AmMaZeKaSc24,
  BaAmCoReKa23, AmCoYeMaHaFaKa23}, verifying object-detection
systems~\cite{ElElIsDuGaPoBoCoKa24}, and systems in the aerospace
domain~\cite{MaAmWuDaNeRaMeDuGaShKaBa24, MaAmWuDaNeRaMeDuHoKaBa24};
and it has also been used as a backend in the formal pruning of neural
networks~\cite{LaKa21}. However, our technique is not bound to a
particular DNN verifier, and other tools could be used as well ---
provided they generate concrete inputs for satisfiable queries.

\mysubsection{Marabou Queries.} In our implementation, properties were defined using runner-ups only, i.e. in the $\epsilon$-environment of each sample $\mathbf{x}$, the property $f(\mathbf{x})_{i}>f(\mathbf{x})_{\hat{y}}$ was only defined for the class of the model's runner-up and not all of the classes. The $\epsilon$ values chosen were the ones calculated by FGSM with binary search or by DFAL, increased by a small margin of 0.05 to allow more slack for adversarial examples to be found. For \texttt{BADGE} and \texttt{Random}, where the $\epsilon$ value is not calculated as part of the method, a fixed value of $\epsilon=0.01$ was chosen. The runtime limit for each query is 30 minutes.
\clearpage

\section{Algorithms} \label{app:algorithms}

\vspace{-0.7cm}

\begin{algorithm}[ht]
\caption{\textsc{FGSM with Binary Search for Margin Estimation}}
\label{alg:fgsm-binary-updated}
\begin{algorithmic}[1]
\Require Network $f$, input $x$, tolerance $\tau$
\Ensure Estimated boundary perturbation $\epsilon^*$ and adversarial $x^*$
\State $start \gets 0$, $end \gets 1$, $\epsilon \gets \tfrac{1}{2}$
\While{$end - start > \tau$}
  \State $x_{\text{adv}} \gets \text{FGSM}(f, x, \epsilon)$ 
  \If{$\arg\max f(x_{\text{adv}}) \neq \arg\max f(x)$}
     \State $end \gets \epsilon$
  \Else
     \State $start \gets \epsilon$
  \EndIf
  \State $\epsilon \gets start + \frac{end - start}{2}$
\EndWhile
\State $\epsilon^* \gets \epsilon$; \ $x^* \gets \text{FGSM}(f, x, \epsilon^*)$
\State \Return $\epsilon^*, x^*$
\end{algorithmic}
\end{algorithm}

\vspace{-0.9cm}

\begin{algorithm}[H]
\caption{\textsc{Formal Verification-Based Adversarial Active Learning (FVAAL)}}
\label{alg:fvaal}
\begin{algorithmic}[1]
\Require Network $f_\theta$, verifier $\mathcal{V}$, unlabeled pool $\mathcal{U}$, oracle $\mathcal{O}$,
         number of initial training samples $N_{\text{init}}$, number of iterations $T$, size of data pool sub-sampled each iteration $N_{\text{sub}}$, number of samples to query each iteration $N_{\text{query}}$, number of adversarial examples to generate $N_{\text{adv}}$, tolerance $\tau$
\Ensure Final model $f_\theta^{(T)}$
\State $\mathcal{L} \gets$ $N_{\text{init}}$ random samples from $\mathcal{U}$ with labels via $\mathcal{O}$; \quad $\mathcal{U}\gets \mathcal{U}\setminus \mathcal{L}$
\State \textbf{Train} $f_\theta^{(0)}$ on $\mathcal{L}$
\For{$t=1$ to $T$}
  \State $\mathcal{L}_t \gets \emptyset$
  \State $\mathcal{U}_t \gets$ sample $N_{\text{sub}}$ points from $\mathcal{U}$
  \For{each $x_i\in\mathcal{U}_t$}
     \State $\hat\epsilon_i \gets \textsc{BinarySearchFGSM}(f_\theta^{(t-1)},x_i,\tau)$ \Comment{estimated margin}
  \EndFor
  \State $S \gets$ $N_{\text{query}}$ elements of $\mathcal{U}_t$ with smallest $\hat\epsilon_i$
  \State Obtain labels $y_i\gets \mathcal{O}(x_i)$ for $x_i\in S$; \quad $\mathcal{L}_t \gets \mathcal{L}_t \cup \{(x_i,y_i): x_i\in S\}$
  \For{each $(x_i,y_i)\in \mathcal{L}_t$}
     \State $A_i \gets$ $N_{adv}$ adversarial examples generated by $\mathcal{V}$
     \State $\mathcal{L}_t \gets \mathcal{L}_t \cup \{(x',y_i): x'\in A_i\}$
  \EndFor
  \State $\mathcal{L} \gets \mathcal{L} \cup \mathcal{L}_t$; \quad $\mathcal{U}\gets \mathcal{U}\setminus S$
  \State \textbf{Train} $f_\theta^{(t)}$ on $\mathcal{L}$ 
\EndFor
\State \Return $f_\theta^{(T)}$
\end{algorithmic}
\end{algorithm}
\clearpage

\section{Selection Criterion}
\label{app:selection-criterion}
\begin{figure}[t]
  \centering

  \subfloat[MNIST]{
    \includegraphics[width=0.31\textwidth]{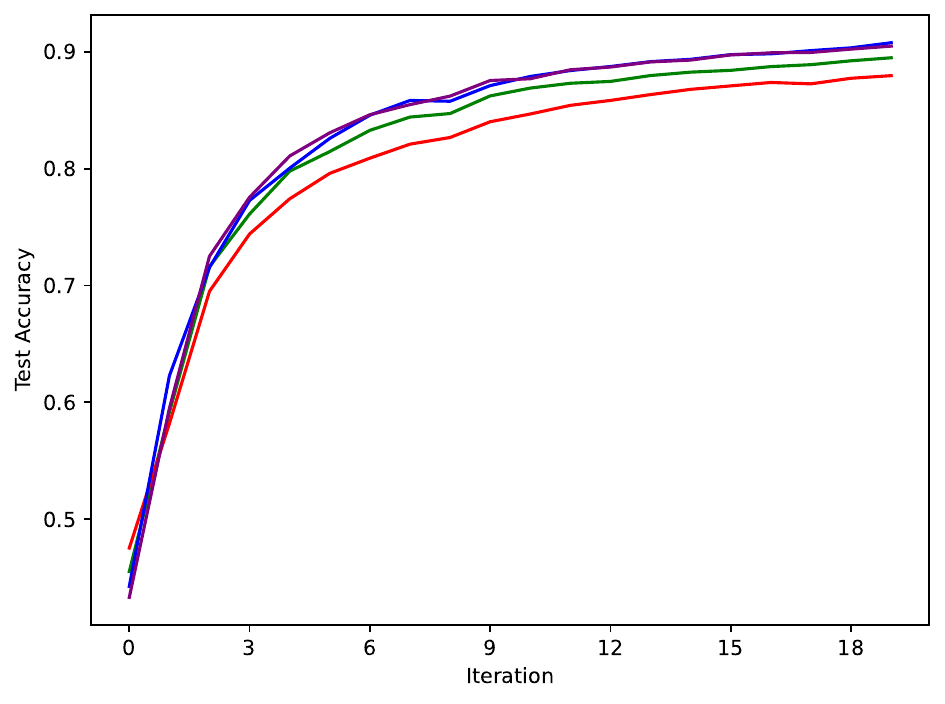}
  }
  \subfloat[fashionMNIST]{
    \includegraphics[width=0.31\textwidth]{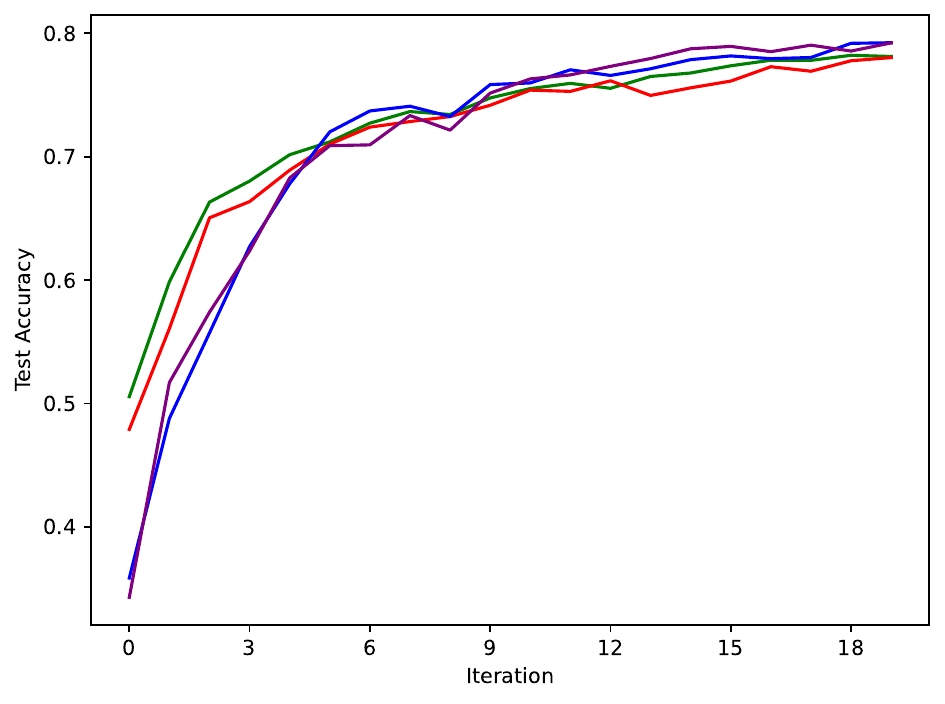}
  }
  \subfloat[CIFAR-10]{
    \includegraphics[width=0.31\textwidth]{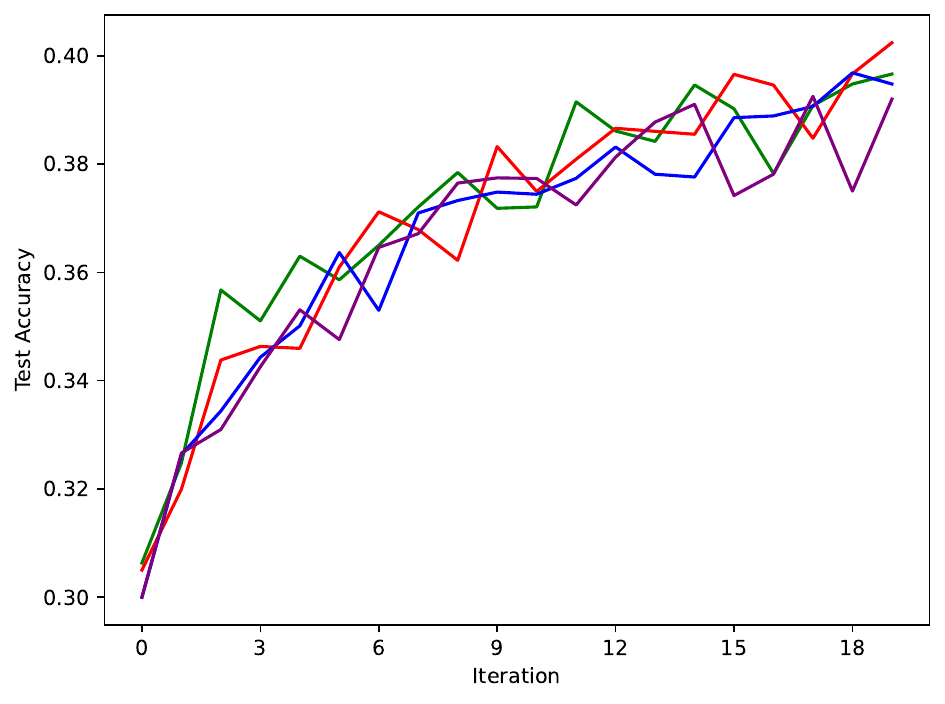}
  }

  \begin{minipage}{0.95\linewidth}\centering\small
  \legenditem{mygreen}{BADGE}\quad
  \legenditem{purple}{DFAL-No-Adv}\quad
  \legenditem{myblue}{FVAAL-No-Adv}\quad
  \legenditem{myred}{Random}\\[0.3ex]
  \end{minipage}

  \caption{Test accuracy vs. number of iterations on the three benchmarks without adding adversarial inputs}
  \label{fig:accuracy-noadv}
\end{figure}

We evaluate \texttt{FVAAL}'s choice of informative samples by conducting an experiment where 
adversarial examples are not added at all to the training set: \texttt{BADGE} and \texttt{Random} are used without modification, 
while for \texttt{DFAL} and \texttt{FVAAL} we use variants where all adversarial inputs are discarded (``\texttt{-No-Adv}''). 
Results are shown in Fig.~\ref{fig:accuracy-noadv}. On MNIST and
fashionMNIST, \texttt{FVAAL-No-Adv} maintains high accuracy, outperforms \texttt{BADGE} and \texttt{Random}
and gives similar results to \texttt{DFAL-No-Adv} at almost every iteration. On CIFAR-10,
the results are noisy, and the various methods appear roughly
equivalent.

\section{Diversity of Adversarial Examples}
\label{app:diversity}
To quantify the diversity of the adversarial samples generated by the \texttt{+FV-Adv} 
and \texttt{+FGSM-Adv} variants, we measure the dispersion of their embeddings in the 
model's feature space. Specifically, for each variant, we consider 50 samples selected 
in the final DAL round according to the respective acquisition strategy. For MNIST and 
FashionMNIST, this set coincides with all selected samples; for CIFAR-10, 50 samples 
were drawn uniformly at random from the selected subset. For each of these inputs, we 
collect all corresponding adversarial examples, extract their penultimate-layer 
embeddings using the trained model of that variant, and compute all pairwise Euclidean 
distances between these feature vectors. Let $d(x_i, x_j)$ denote the distance between 
the embeddings of $x_i$ and $x_j$. Each method is therefore represented by the empirical 
distribution $\{ d(x_i, x_j) \mid i < j \}$.

We summarize these distributions by reporting their mean and standard deviation. This 
analysis is performed across all five independent runs of each benchmark. The reported 
across-run means and standard deviations are aggregated using the law of total variance. 
Table~\ref{tab:feature-diversity} presents these statistics for all variants.

Across nearly all settings, adversarial examples produced via formal verification exhibit 
consistently larger average pairwise distances compared to those generated by FGSM, indicating 
broader coverage of distinct feature-space directions. This suggests that formal verification yields a 
more diverse set of adversarial inputs, potentially reducing redundancy in DAL sampling and improving label efficiency.

\begin{table}[t]
  \centering
  \begin{threeparttable}
  \caption{Feature-space diversity of adversarial examples.
    Mean and standard deviation of all pairwise distances among $50$
    adversarial examples per method, computed in the penultimate-layer
    embedding space. Higher mean indicates greater diversity.}
  \label{tab:feature-diversity}
  \setlength{\tabcolsep}{7pt}
  \begin{tabular}{
      l
      S[table-format=1.3(3)]
      S[table-format=1.3(3)]
      S[table-format=1.3(3)]
      S[table-format=1.3(3)]
      S[table-format=1.3(3)]
  }
    \toprule
    {Method} & {MNIST} & {FashionMNIST} & {CIFAR-10} \\
    \midrule
    BADGE+FGSM-Adv & {$26.186\pm8.848$} & {$28.607\pm14.413$} & {$6.924\pm3.213$} \\
    BADGE+FV-Adv & {$\second{33.631}\pm11.617$} & {$\second{35.0126}\pm15.501$} & {$\best{10.224}\pm4.585$} \\
    DFAL+FGSM-Adv & {$24.468\pm9.077$} & {$27.809\pm16.496$} & {$6.669\pm3.447$} \\
    DFAL+FV-Adv & {$31.412\pm11.932$} & {${32.552}\pm17.388$} & {${8.744}\pm4.074$} \\ 
    FVAAL & {${31.987}\pm11.175$} & {$30.797\pm15.282$} & {$8.658\pm4.141$} \\ 
    Random+FGSM-Adv & {$26.507\pm9.225$} & {$28.666\pm14.709$} & {$7.680\pm3.759$} \\
    Random+FV-Adv & {$\best{36.578}\pm13.228$} & {$\best{36.496}\pm16.862$} & {$\second{9.860}\pm4.746$} \\
    \bottomrule
  \end{tabular}
  \begin{tablenotes}[flushleft]
    \footnotesize
    \item Bold = best per benchmark; \underline{underline} = second best.
  \end{tablenotes}
  \end{threeparttable}
\end{table}

\end{document}